\documentclass{article}

\usepackage[square,numbers]{natbib}
\usepackage[preprint]{neurips_2021}

\usepackage[utf8,latin1]{inputenc} % allow utf-8 input
\usepackage[scaled=0.90]{inconsolata}
\usepackage[T1]{fontenc}    % use 8-bit T1 fonts
\usepackage{hyperref}       % hyperlinks
\usepackage{url}            % simple URL typesetting
\usepackage{booktabs}       % professional-quality tables
\usepackage{amsfonts}       % blackboard math symbols
\usepackage{nicefrac}       % compact symbols for 1/2, etc.
\usepackage{microtype}      % microtypography
\usepackage{xcolor}         % colors
\usepackage{multirow}

\usepackage{graphicx} % more modern
\usepackage{amsmath,amssymb,amstext}
\usepackage{xspace}
\usepackage{amsmath,amsfonts,bm}

\def\eqref#1{equation~\ref{#1}}
\def\1{\bm{1}}

\def\vb{{\bm{b}}}

\def\vs{{\bm{s}}}

\def\vv{{\bm{v}}}

\def\mV{{\bm{V}}}

\DeclareMathAlphabet{\mathsfit}{\encodingdefault}{\sfdefault}{m}{sl}
\SetMathAlphabet{\mathsfit}{bold}{\encodingdefault}{\sfdefault}{bx}{n}

\newcommand{\R}{\mathbb{R}}

\newcommand{\mr}[2]{\multirow{#1}{*}{#2}}
\newcommand{\mc}[3]{\multicolumn{#1}{#2}{#3}}

\usepackage{macros}

\title{Rotation Invariant Graph Neural Networks using Spin Convolutions}

\author{
  Muhammed Shuaibi$^1$, Adeesh Kolluru$^1$, Abhishek Das$^2$, Aditya Grover$^2$, \\
  \textbf{Anuroop Sriram}$^2$, \textbf{Zachary Ulissi}$^1$, \textbf{C. Lawrence Zitnick}$^2$ \\
  $^1$ Carnegie Mellon University\\
  $^2$ Facebook AI Research
}

\begin{document}

\maketitle

\begin{abstract}
Progress towards the energy breakthroughs needed to combat climate change can be significantly accelerated through the efficient simulation of atomic systems. Simulation techniques based on first principles, such as Density Functional Theory (DFT), are limited in their practical use due to their high computational expense. Machine learning approaches have the potential to approximate DFT in a computationally efficient manner, which could dramatically increase the impact of computational simulations on real-world problems.

Approximating DFT poses several challenges. These include accurately modeling the subtle changes in the relative positions and angles between atoms, and enforcing constraints such as rotation invariance or energy conservation. We introduce a novel approach to modeling angular information between sets of neighboring atoms in a graph neural network. Rotation invariance is achieved for the network's edge messages through the use of a per-edge local coordinate frame and a novel spin convolution over the remaining degree of freedom. Two model variants are proposed for the applications of structure relaxation and molecular dynamics. State-of-the-art results are demonstrated on the large-scale Open Catalyst 2020 dataset. Comparisons are also performed on the MD17 and QM9 datasets.

\end{abstract}

\section{Introduction}
\label{sec:intro}
Many of the world's challenges such as finding energy solutions to address climate change~\cite{zitnick2020introduction, OC20} and drug discovery~\cite{ramakrishnan2014quantum,senior2020improved} are fundamentally problems of atomic-scale design. A notable example is the discovery of new catalyst materials to drive chemical reactions that are essential for addressing energy scarcity, renewable energy storage, and more broadly climate change~\cite{zitnick2020introduction, rolnick2019tackling}. Potential catalyst materials are typically modeled using Density Functional Theory (DFT) that estimates the forces that are exerted on each atom and the energy of a system or structure of atoms. Unfortunately, the computational complexity of DFT limits the scale at which it can be applied. Efficient machine learning approximations to DFT calculations hold the potential to significantly increase the discovery rate of new materials for these important global problems.

Graph Neural Networks (GNNs) \cite{gori2005new,zhou2020graph} are a common approach to modeling atomic structures, where each node represents an atom and the edges represent the atom's neighbors \cite{schutt2017quantum,gilmer2017neural,jorgensen2018neural,schutt2017schnet,schutt2018schnet,xie2018crystal,qiao2020orbnet,klicpera2020directional}. A significant challenge in designing models is utilizing relative angular information between atoms, while maintaining a model's invariance to system rotations. Numerous approaches have been proposed, such as only using the distance between atoms \cite{schutt2017schnet,schutt2018schnet,xie2018crystal}, or limiting equivariant angular representations to linear transformations to maintain equivariance \cite{weiler20183d, batzner2021se, anderson2019cormorant,thomas2018tensor}. One promising approach is the use of triplets of neighboring atoms to define local coordinate frames that are invariant to system rotations \cite{klicpera2020directional,klicpera_dimenetpp_2020}. The relative angles between the three atoms may be used to update the GNN's messages while maintaining the network's invariance to rotations. It has been shown that this additional angular information results in significantly improved accuracies on several tasks \cite{klicpera2020directional,klicpera_dimenetpp_2020,OC20}.

\begin{figure}
\centering
\includegraphics[width=0.95\linewidth]{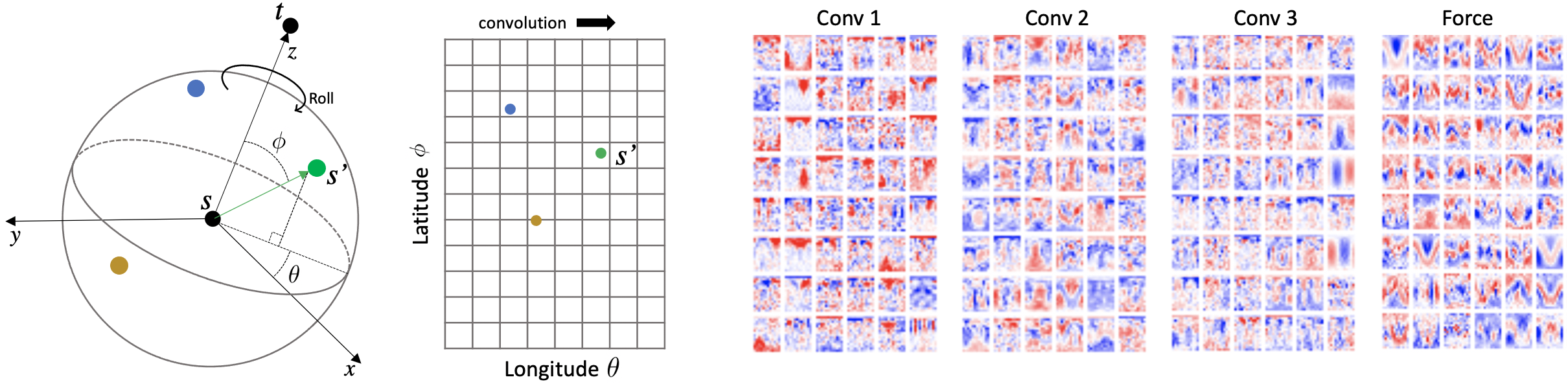}
\caption{Illustration of projecting an atom $\acute{s}$ in the neighborhood of $s$ onto a sphere in a local coordinate frame defined by atom $s$ and $t$ (left). For each projected atom, a corresponding latitude $\phi$ (inclination) and longitude $\theta$ (azimuth) is computed for its projection onto a 2D reference frame (middle). The spin convolution is done in the longitudinal direction, corresponding to a roll is 3D space. (right) Example channel filters that are learned using the grid-based approach for the first through third message blocks and the force block.}
\label{fig:sphere}
\end{figure}

We propose encoding angular information using a local reference frame defined by only two atoms; the source and target atoms for each edge in a GNN. Using this reference frame, a spherical representation of the incoming messages to the source atom is created, Figure \ref{fig:sphere}. The representation has the benefit of encoding all neighboring atom information, and not just information between atom triplets, which may result in higher-order information being captured. The complication is a reference frame defined by two atoms (or two 3D points) still has one remaining degree of freedom - the roll rotation about the axis defined by the two 3D points. If this final degree of freedom is not accounted for, the model will not be invariant to system rotations. Our solution is to perform a convolution on the spherical representation across this final rotation, called a ``spin convolution''. By globally pooling the convolution's features, the resulting \model~model maintains rotation invariance while enabling the capture of rich angular information. 

We describe two model variations that are used depending on the importance of energy conservation in the final application. We propose an energy-centric model that enforces energy conservation by calculating the forces using the negative partial derivative of the energy with respect to the atoms' positions \cite{chmiela2017machine}. Our second approach is a force-centric model that directly estimates the atom forces that is not energy conserving. While the force-centric model's energy estimation is rotation invariant, the model's final force estimation layer is not strictly rotation equivariant, but through its architectural design it is encouraged to learn rotation equivariance during training. 

Results are demonstrated on the Open Catalyst 2020 (\ocd) dataset \cite{OC20} aimed at simulating catalyst materials that are useful for climate change related applications. The \ocd~dataset contains over 130M training examples for approximating the DFT-estimated forces and energies. Our \model~model achieves state-of-the-art performance for both energy and force estimation. Notably, the force-centric variant, which is not energy conserving, outperforms the energy-centric models. Significant gains in accuracy are achieved for predicting relaxed energies from initial structures, by using the force-centric approach to predict the relaxed structure followed by its energy. Ablation studies are performed on numerous architectural choices, such as the choice of spherical representation and the size of the model. For completeness, we also evaluate our model on the MD17 \cite{chmiela2017machine,chmiela2018towards} and QM9 \cite{ramakrishnan2014quantum} datasets that measure accuracy for molecular dynamics and property prediction tasks respectively for small molecules. Results compare favorably with respect to state-of-the-art methods.

\section{Approach}
\label{sec:approach}

We model a system or structure of atoms using a Graph Neural Network (GNN) \cite{gori2005new,li2015gated,zhou2020graph}, where the nodes represent atoms and the edges represent the atoms' neighbors. In this section, we describe both an energy-centric and force-centric model to estimating atomic forces, which vary in how they estimate forces and whether they are energy conserving. We begin by describing the components shared by each approach, followed by how these components are used. Code will be released upon acceptance under a permissive open-source license.

\subsection{Inputs and Outputs}

The inputs to the network are the 3D positions $\bm{x}_i$ and the atomic numbers $a_i$ for all $i \in n$ atoms. The outputs are the per atom forces $\bm{f}_i \in \R^3$ and the overall structure's energy $E$.
The 3D distance offset between a pair of source and target atoms $s$ and $t$ respectively is $\bm{x}_{st} = \bm{x}_s - \bm{x}_t$ with a distance of $d_{st} = \|\bm{x}_{st}\|_2$. Directional information is encoded using the normalized unit vector $\bm{\hat{x}}_{st} = \bm{x}_{st} / d_{st}$. 

The graph neural network is constructed with each atom $t$ as a node and the edges representing the atom's neighbors $s \in N_t$, where $N_t$ contains all atoms $s$ with $d_{st} < \delta$.  Each edge has a corresponding message $m_{st}$ that passes information from atom $s$ to $t$. The output forces and energy are computed as a function of edge messages $m_{st}$ that we describe next.

\subsection{Energy and force estimation}

The energy-centric and force-centric models compute the structure's energy $E$ as an output. Our GNN model updates for each edge an $M$-dimensional hidden message $\bm{h}^{(k)}_{st} \in \R^M$ for $K$ iterations. The structure's energy $E\in\R$ is computed as a function of the final layer of the edge messages in the GNN:
\begin{equation}
E(\bm{x}, a) = \sum_t \bm{F}_e(a_t, \sum_s \bm{h}^{(K)}_{st}),
\label{eqn:energy}
\end{equation}
where $\bm{F}_e$ is a single embedding block described later. As we also discuss later, the edge messages $\bm{h}_{st}$ are invariant to system rotations, so the estimated energy $E$ is also invariant. 

The estimation of the forces varies for the energy-centric and force-centric models. The energy-centric model estimates the forces using the negative partial derivative of the energy with respect to the atom positions. This approach to force estimation has the benefit of enforcing energy conservation \cite{chmiela2017machine}, i.e., the forces along any closed path sum to zero. The calculation of the partial derivative \cite{chmiela2017machine,schutt2017schnet,schutt2018schnet} requires an additional step similar to performing backpropagation when updating the network's weights:
\begin{equation}
\bm{f} = -\frac{\partial}{\partial \bm{x}}E(\bm{x}, a)
\end{equation}

The force-centric model estimates forces directly for an atom $t$ using:
\begin{equation}
\bm{f}_t = \bm{F}_f(a_t, \hat{\bm{x}}_{t}, \bm{h}^{(K)}_{t}),
\label{eqn:force}
\end{equation}
where $\bm{F}_f$ is the force block we describe later, $\hat{\bm{x}}_{t}$ are all the normalized unit vectors for the neighbors of $t$ and $\bm{h}^{(K)}_{t}$ are all incoming messages to atom $t$. This has the benefit of improved efficiency since it does not require an extra backward pass to estimate the forces. The tradeoff is that it does not enforce energy conservation, i.e., the sum of the forces along a closed path may not equal zero. Depending on the application, an energy-centric or force-centric approach may be most suitable. In either model, losses may be applied to both the energy and force estimates with weights determined by the needs of the application.

\begin{figure}
\centering
\includegraphics[width=0.95\linewidth]{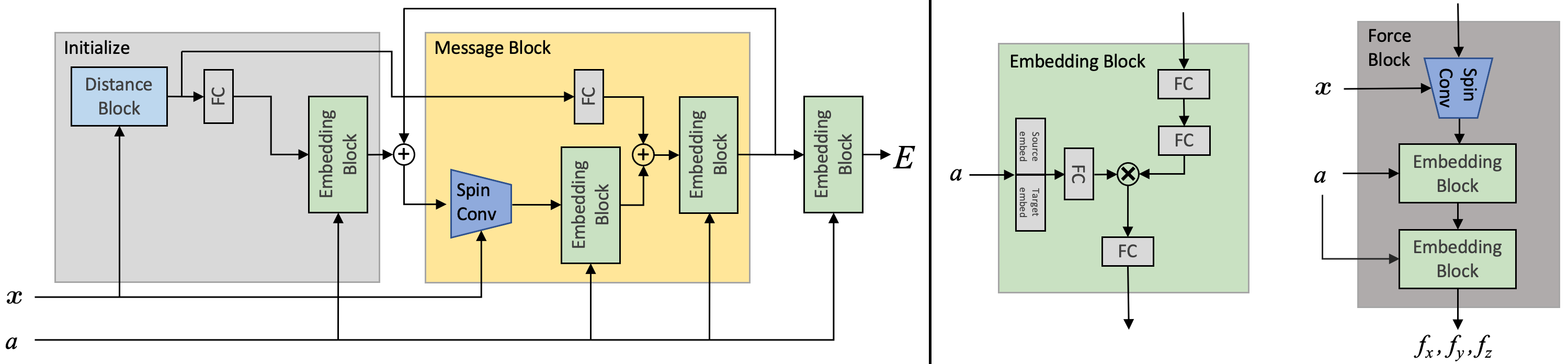}
\caption{(left) Overall model diagram for energy-centric model taking atom positions $\bm{x}$ and atomic numbers $a$ as input and estimating the energy $E$. (right) Diagram of the embedding and force blocks. The force block is only used in the force-centric model to estimate the per-atom forces after the message blocks.}
\label{fig:model}
\end{figure}

\subsection{Messages}

The edge messages are iteratively updated to allow information from increasingly distant atoms to be captured. Each message is represented by a tuple, $m_{st} = \{\hat{\bm{x}}_{st}, d_{st}, \bm{h}^k_{st}\}$, where $\bm{h}^k_{st}$ is the message's hidden state at iteration $k$. Both $\bm{\hat{x}}_{st}$ and $d_{st}$ are used to update the message's hidden state $\bm{h}_{st}$, which is itself rotation invariant due to the spin convolution that we describe later. The hidden state $\bm{h}_{st} \in \R^M$ is updated using:
\begin{equation}
    \bm{h}_{st}^{(k+1)} = \bm{h}_{st}^{(k)} + \bm{F}_h\left(a_s, a_t, m_{st}^{(k)}, m_s^{(k)}\right),
\end{equation}
where $m_s^{(k)}$  is the set of messages coming into node $s$, i.e., all $m_{\acute{s}s}$ with $\acute{s} \in N_s$. The form of $\bm{F}_h$ is illustrated in Figure \ref{fig:model}. It contains three parts; the spin convolution that transforms a spherical projection of the messages into a rotation invariant representation, the distance block that encodes the distance $d_{st}$ between atoms, and the embedding block that incorporates information about the atoms' atomic numbers. The output of the spin convolution is passed through an embedding block, added to the output of the distance block and finally passed through another embedding block. We describe each of these parts in turn. The hidden messages are initialized using just a distance block followed by and embedding block, Figure \ref{fig:model}.

\subsubsection{Spin Convolution}

The spin convolution captures information about the neighbors $\acute{s}\in N_s$ of atom $s$ when updating the message hidden state $\bm{h}_{st}$. The spin convolution has three stages that we describe in turn; projection, convolution and pooling. The convolution captures the relative angular information between the neighboring atoms, and the pooling ensures the output $D$-dimensional feature representation is invariant to system rotations. 

An important feature is the angular information of the neighboring atoms in $N_s$ relative to $s$ and $t$. This information is encoded by creating a local reference frame in which atom $s$ is the center $(0,0,0)$ and the z-axis  points from atom $s$ to atom $t$. As shown in Figure \ref{fig:sphere}(left), this fixes all degrees of freedom except the roll rotation about the vector from $s$ to $t$. The spin convolution is performed across a discretized set of rotations about the roll rotation axis. At each rotation, the atoms $\acute{s}$ are projected onto a sphere centered on $s$ and used to create a spherical representation of the hidden states $\bm{h}_{\acute{s}s}$. Each atom $\acute{s} \in N_s$ is projected using a polar coordinate frame $(\phi, \theta)$ where $\phi$ may be viewed as the latitude (inclination) and $\theta$ as the longitude (azimuth). The polar coordinates are computed in the local edge coordinate frame using $\bar{\bm{x}}_{\acute{s}s} = \mathbf{R}_{st}\hat{\bm{x}}_{\acute{s}s}$ where $\mathbf{R}_{st}$ is a 3D rotation matrix that satisfies $\mathbf{R}_{st}\hat{\bm{x}}_{st} = (0, 0, 1)$. To capture the rich information encoded in the relative angular information between atoms, a set of filters is applied to the spherical representation (Figure \ref{fig:sphere}(right)), similar to how a filter is applied to an image patch with traditional CNNs. 

We explore two potential spherical representations: spherical harmonics and a grid-based approach. Spherical harmonics represent a spherical function using a set of basis functions that are equivariant to rotations. The degree $\ell$ indicates the number of basis functions $L = (\ell + 1)^2$ used. The spherical representation of the incoming messages for each atom is $\R^L \times \R^M$, where $M$ is the size of the message hidden states in $\bm{h}$. The second approach uses the computed polar coordinates $(\phi, \theta)$ for all $\acute{s} \in N_s$ to create a grid-based representation, Figure \ref{fig:sphere}(middle). The polar coordinates are discretized creating a $\R^{\phi} \times \R^{\theta} \times \R^M$ feature representation. Each message hidden state $\bm{h}_{s\acute{s}}^{(k)} \in \R^M$ is added to the 3D feature representation using bilinear interpolation with its corresponding $(\phi, \theta)$.

A 1D convolution is performed with either spherical representation in the longitudinal direction. Filters have the same size as the feature representation, $\R^L \times \R^M$ or $\R^{\phi} \times \R^{\theta} \times \R^M$ for spherical harmonics and the grid-based approach respectively. Full coverage filters are used since the angular relationship between atoms at distant angles is important, e.g., the forces of atoms at exactly $180^{\circ}$ from each other may cancel out. Large filters also enable the network to learn the complex relationships between numerous neighboring atoms. Rotations are performed using Wigner D-matrices for the spherical harmonic representation, while a simple translation is used for the grid-based representation. The result of the convolution is a $\R^\theta \times \R^D$ feature vector corresponding to $D$ filters applied to each longitudinal orientation. To make the representation invariant to rotations, average pooling is performed in the longitudinal direction resulting in a final $\R^D$ feature vector.

%The latitude $\phi$ for each neighboring atom $\acute{s}$ to $s$ is invariant to system rotations, but the longitude $\theta$ is not. However, differences in longitude are invariant, since all points rotate uniformly with system rotations. Thus we want a representation that captures the relative differences in latitude and longitude between neighboring atoms, but not their absolute longitude.

\subsubsection{Distance Block}

The distance block encodes the distance between two atoms. The distance is encoded using a set of evenly distributed Gaussian basis functions $\mathcal{G}$ with means $\mu_i$ and standard deviation $\sigma$. The means of the basis functions are evenly distributed from 0 to $\delta$ angstroms. Since the atomic radii of each element varies, the relative position of two atoms $s$ and $t$ is highly dependent on their atomic numbers $a_s$ and $a_t$. To account for this, gain $v_{a_s a_t}$ and offset $u_{a_s a_t}$ scalars for the distance $d_{st}$ are learned for each potential pair of atomic numbers:
\begin{equation}
\vb_i = \mathcal{G}_i(v_{a_s a_t}d_{st} + u_{a_s a_t} - \mu_i, \sigma)
\end{equation}
The resulting feature $\vb$ is passed through a linear transformation to create a $D$-dimensional feature vector that is passed to the next block.

\begin{figure}
\centering
\includegraphics[width=0.998\linewidth]{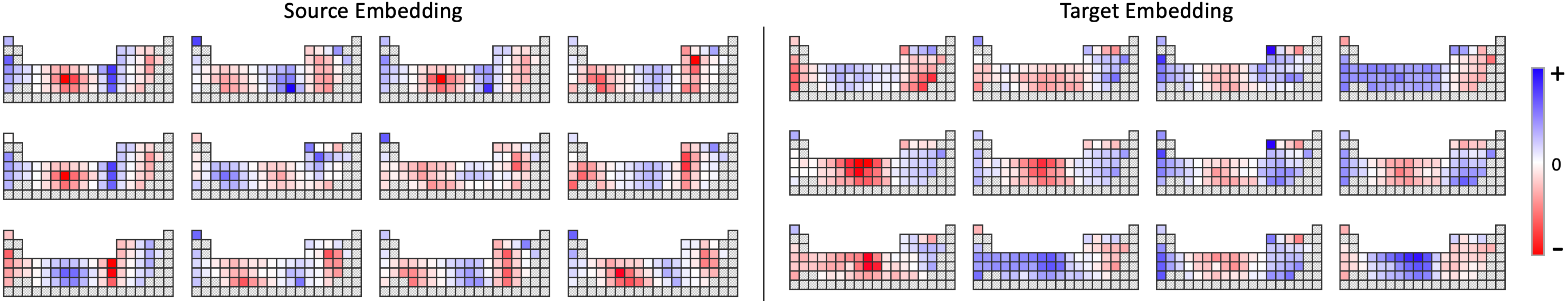}
\caption{Illustration of learned embeddings (weights on the one-hot embeddings) for the source $a_s$ and target $a_t$ atomic numbers plotted on a periodic table. A random sample of 12 values from each embedding are shown. Embeddings are from the first embedding block in the first message update. Note that neighboring atoms in the periodic table with similar properties have similar weights. Elements not in the OC20 dataset are marked with a light grey checkerboard pattern.}
\label{fig:embedding}
\end{figure}

\subsubsection{Embedding Block}

The embedding block incorporates the atomic number information $a_s$ and $a_t$ into the update of the message's hidden state. The embedding operation may be interpreted as a mixture of experts \cite{masoudnia2014mixture} approach that computes $B$ different variations of the input, which are weighted by an embedding computed from the atoms' atomic numbers. The block's inputs are used to compute $B$ sets of hidden values $\mV_{st} \in \R^D \times \R^B$. A one-hot embedding for the atomic numbers $a_s$ and $a_t$ are concatenated and used to compute an $B$ dimensional vector, $\vv_{st} \in \R^B$, for weighting the $B$ different sets of hidden values. An illustration of the learned embeddings are shown in Figure \ref{fig:embedding}. $\vv_{st}$ is computed using a two layer network and softmax. The matrix $\mV_{st}$ is multiplied by vector $\vv_{st}$ resulting in a vector of length $D$. As shown in Figure \ref{fig:model}, the result is passed through an additional fully connected layer before being passed to the next block. The output of the block is either $D$ if it is used in the message update. If the embedding block is used to compute the final energy, only the atomic number $a_t$ embedding is used, the input dimension is $M$ instead of $D$, and the output is size 1.

\subsection{Force Block}

The force block computes the per-atom 3D forces $f$ from $a_t$, $\hat{\bm{x}}_{t}$, and $\bm{h}^{(K)}_{t}$ using Equation (\ref{eqn:force}). The force block uses a similar spin convolution as the message block, except the sphere is centered on the target atom $t$ and is orientated along the $x, y$ and $z$ axes to compute $f_x, f_y$ and $f_z$ respectively. That is, the force block is used three times to compute the force magnitude in each orthogonal direction for each atom. The force block uses the same embedding blocks as message passing, Figure \ref{fig:model}. 

The same weights are used to compute forces in each of the three directions, only the orientation of the sphere used to create the convolutional features changes. To add more robustness to the force estimation and encourage rotational equivariance, the overall structure may be randomly rotated several times and the forces estimated. The multiple estimates may then be rotated back to the original reference frame and averaged. For both training and testing, five random rotations are used. Empirically, this approach encourages the networks to learn an approximate rotation equivariant representation even though rotation equivariance is not strictly enforced.

\section{Experiments}
\label{sec:experiments}
In this section, we begin by presenting our primary results on the Open Catalyst 2020 (OC20) dataset \cite{OC20} and compare against state-of-the-art models. This is followed by results on the smaller datasets of MD17 \cite{chmiela2017machine,chmiela2018towards} and QM9 \cite{ramakrishnan2014quantum} for additional model comparison.

\begin{table*}[t]
    \centering

    \renewcommand{\arraystretch}{1.0}
    \setlength{\tabcolsep}{5pt}
    \resizebox{0.97\linewidth}{!}{
    \begin{tabular}{lrrrrrccccc}
      \toprule
         \mr{2}{\textbf{Model}} & \textbf{Hidden} & \textbf{\#Msg} & \mr{2}{\textbf{\#Params}} & \textbf{Train}  & \textbf{Inference} & \mc{4}{c}{\textbf{\ocd{} Test}} \\
        & \textbf{dim} & \textbf{layers} & & \textbf{time} & \textbf{time} & Energy MAE [eV] $\downarrow$ & Force MAE [eV/\AA] $\downarrow$ & Force Cos $\uparrow$ & EFwT [\%] $\uparrow$  \\
      \midrule
        Median & -- & -- & -- & & & 2.258 & 0.08438 & 0.0156 & 0.005 \\
      \midrule
        SchNet\cite{schutt2018schnet,OC20}  &  1024 & 5 & 9.1M & 194d & 0.8h &  -- & 0.04903	& 0.3413 & 0   \\
        DimeNet++\cite{klicpera_dimenetpp_2020,OC20} & 192 & 3 & 1.8M &  587d & 8.5h & 0.5343  & 0.04758 & 0.3560 & 0.05  \\
        DimeNet++ energy-only\cite{klicpera_dimenetpp_2020, OC20} & 192 & 3 & 1.8M &  587d & 8.5h & 0.4802 & 0.3459 & 0.1021 & 0.0  \\
        DimeNet++ force-only\cite{klicpera_dimenetpp_2020, OC20} & 192 & 3 & 1.8M &  587d & 8.5h & -- & 0.03573 & 0.4785 &	-- \\
        DimeNet++-large\cite{klicpera_dimenetpp_2020, OC20}  & 512 & 3 & 10.7M & 1600d & 27.0h  & --  & 0.03275 & \textbf{0.5408} & -- \\
        ForceNet\cite{hu2021forcenet} & 512 & 5 & 11.3M & 31d & 1.3h &  -- & 0.03432 & 0.4770 & --  \\
        ForceNet-large\cite{hu2021forcenet} & 768 & 7 & 34.8M & 194d & 3.5h &  - & 0.03113 & 0.5195 & - \\
        \midrule
        {\bf \model{} (energy-centric)}  & 256 & 3 & 6.1M & 275d & 22.7h & 0.4114 & 0.03888 & 0.4299  & 0.16  \\
        {\bf \model{} (energy-centric) force-only }  & 256 & 3 & 6.1M & 380d & 22.7h & -- & 0.03258 & 0.4976 & -- \\
        {\bf \model{} (force-centric)}  & 256 & 3 & 8.5M & 275d & 9.1h & \textbf{0.3363} & \textbf{0.02966} & 0.5391 & \textbf{0.45}  \\
      \bottomrule
    \end{tabular}
    }
    \caption{Comparison of \model{} to existing GNN models on the S2EF task. Average results across all four test splits are reported. We mark as bold the best performance and close ones, \ie, within 0.0005 MAE, which according to our preliminary experiments, is a good threshold to meaningfully distinguish model performance. Training time is in GPU days, and inference time is in GPU hours. Median represents the trivial baseline of always predicting the median training force across all the validation atoms.}
    \vspace{-0.35cm}
    \label{tab:comp-s2ef}
\end{table*}

\begin{table*}[t]
    \centering

    \renewcommand{\arraystretch}{1.0}
    \setlength{\tabcolsep}{5pt}
    \resizebox{0.96\linewidth}{!}{
    \begin{tabular}{lcccccccc}
      \toprule
         \mr{2}{\textbf{Model}} & \mc{4}{c}{\textbf{Energy MAE (eV)} $\downarrow$} &  \mc{4}{c}{\textbf{Force MAE (eV/\AA)} $\downarrow$} \\
        & ID & OOD Ads. & OOD Cat. & OOD Both & ID & OOD Ads. & OOD Cat. & OOD Both \\
      \midrule
        Median & 2.043 & 2.420 & 1.992 & 2.577 & 0.0809 & 0.0801 & 0.0787 & 0.0978 \\
      \midrule
      & \mc{8}{c}{Energy Loss Only} \\
        SchNet  & 0.395 & 0.446 & 0.551 & 0.703 & - & - & - & - \\
        DimeNet++ & 0.359 & 0.402 & 0.506 & 0.654 & - & - & - & -  \\
      \midrule
      & \mc{8}{c}{Force Loss Only} \\
        SchNet  & - & - & - & - & 0.0443 & 0.0469 & 0.0459 & 0.0590   \\
        DimeNet++ & - & - & - & - & 0.0331 & 0.0341 & 0.0340 & 0.0417 \\
        DimeNet++-large  & - & - & - & - & 0.0281 & 0.0289 & 0.0312 & 0.0371  \\
        ForceNet & - & - & - & - & 0.0313 & 0.0320 & 0.0331 & 0.0409   \\
        ForceNet-large & - & - & - & - & 0.0278 & 0.0283 & 0.0309 & 0.0375  \\
        {\bf \model{} (energy-centric)}  & - &	- &	- &	- &	0.0309&	0.0321&	0.0315&	0.0393   \\
      \midrule
      & \mc{8}{c}{Energy and Force Loss} \\
        SchNet  & 0.443 & 0.491 & 0.529 & 0.716 & 0.0493 & 0.0527 & 0.0508 & 0.0652   \\
        DimeNet++ &  0.486 & 0.470 & 0.533 & 0.648 & 0.0443 & 0.0458 & 0.0444 & 0.0558 \\
        {\bf \model{} (energy-centric)}  & 0.351 &	0.367 &	0.411 & 0.517 & 0.0358 & 0.0374 & 0.0364 & 0.0460    \\
        {\bf \model{} (force-centric)}  & \textbf{0.261} & \textbf{0.275} & \textbf{0.350} & \textbf{0.459} & \textbf{0.0269} & \textbf{0.0277} & \textbf{0.0285} & \textbf{0.0356}  \\
        \bottomrule
    \end{tabular}
    }
    \caption{Comparison of \model{} to existing GNN models on different test splits.
    We mark as bold the best performance and close ones, \ie, within 0.0005 MAE, which according to our preliminary experiments, is a good threshold to meaningfully distinguish model performance. Training time is in GPU days, and inference time is in GPU hours.
    Median represents the trivial baseline of always predicting the median training force across all the validation atoms. }
    \vspace{-0.35cm}
    \label{tab:comp-splits}
\end{table*}

\paragraph{Implementation details.}

For all models, the edge messages have size $M=32$ with $K=3$ layers, the hidden dimension $D=256$ and embedding dimension $B=8$. Unless otherwise stated, the convolutional filters are of size 16x12 and 12x8 for the force-centric and energy-centric models respectively. A smaller filter size was used for the energy-centric model due to memory constraints. GroupNorm \cite{wu2018group} is applied after the spin convolution with group size 4. An L1 loss is used for all experiments. The force loss was weighed by 100 with respect to the energy loss, except for the force-only model where the energy loss is set to 0. All models were trained with Adam (amsgrad) to convergence with the learning rate multiplied by 0.8 when the validation error plateaus. Training was performed using batch sizes ranging from 64 to 96 samples across 32 Volta 32GB GPUs. The Swish \cite{ramachandran2017searching} function is used for all non-linear activation functions. The neighbors $s\in N_t$ of each atom $t$ are found using a distance threshold of $\delta = 6$\AA. If more than 30 atoms are within the distance threshold, only the closest 30 are used.  The distance block uses 256 to 512 Gaussian basis functions with $\sigma$'s equal to three times the distance between Gaussian means. 

\subsection{OC20}

The OC20 dataset \cite{OC20} contains over 130 million structures used to train models for predicting forces and energies during structure relaxations that is released under a CC Attribution 4.0 License. Since the goal of a structure relaxation is to find a local energy minimum, energy conservation in optional for this task. We report results for the Structure to Energy and Forces (S2EF), the Initial Structure to Relaxed Energy (IS2RE) and the Initial Structure to Relaxed Structure (IS2RS) tasks. 

\subsubsection{Structure to Energy and Forces (S2EF)}

There are four metrics for the S2EF task, the energy and force Mean Absolute Error (MAE), the Force Cosine similarity, and the Energy and Forces within a Threshold (EFwT). The EFwT metric is meant to indicate the percentage of energy and force predictions that would be useful in practice. Results for three model variants are shown in Table \ref{tab:comp-s2ef} on the test set. The \model~force-centric approach has the lowest energy MAE and force MAE of all models.  While still low in absolute terms, the \model~models are improving over other models on the EFwT metric. DimeNet++-large slightly out performs \model~on the force cosine metric. The training time for the \model~is significantly faster than DimeNet++, while being a little slower than ForceNet \cite{hu2021forcenet} or SchNet \cite{schutt2018schnet}.

\begin{table*}[t]
    \centering
    \renewcommand{\arraystretch}{1.0}
    \setlength{\tabcolsep}{5pt}
    \resizebox{0.97\linewidth}{!}{
    \begin{tabular}{lrrrrcccc}
      \toprule
         \mr{2}{\textbf{Model}} & \textbf{Hidden} & \textbf{\#Msg} & \mr{2}{\textbf{\#Params}} & \textbf{Train}  &  \mc{4}{c}{\textbf{\ocd{} Val ID 30k}} \\
        & \textbf{dim} & \textbf{layers} & & \textbf{time} & Energy MAE [eV] $\downarrow$ & Force MAE [eV/\AA] $\downarrow$ & Force Cos $\uparrow$ & EFwT [\%] $\uparrow$  \\
      \midrule
        Median & & & & & &   &  & \\
      \midrule 
        \mc{1}{c}{\textbf{Energy-Centric}} & & & & & & & & \\
        {\model{} (grid 12x8)}  & 128 & 2 & 1.3M & 54d &  -- & 0.0417 & 0.401 & -- \\
        {\model{} (spherical harmonics, $\ell = 5$)}  & 256 & 3 & 6.4M & 119d & --  & 0.0405 & 0.411 & -- \\
        {\model{} (grid 12x8)}  & 256 & 3 & 6.1M & 87d &  -- & 0.0406 & 0.426 & -- \\
    \midrule
    \mc{1}{c}{\textbf{Force-Centric}} & & & & & & & & \\
        
        {\model{} (grid 12x8)}  & 128 & 2 & 1.8M & 54d & 0.376 & 0.0370 & 0.436 &  0.15\%   \\
        {\model{} (grid no conv 16x12)}  & 256 & 3 & 8.5M & 56d & 0.341 & 0.0348 &  0.462  & \textbf{0.20}\%    \\
        {\model{} (spherical harmonics, $\ell = 5$)} & 256 & 3 & 8.1M & 113d & 0.321 & \textbf{0.0328} & \textbf{0.484} & \textbf{0.22}\%    \\
        {\model{} (grid 16x12)}  & 256 & 3 & 8.5M & 76d & \textbf{0.317} & \textbf{0.0326} & \textbf{0.484} & \textbf{0.20}\%    \\
      \bottomrule
    \end{tabular}}
    \caption{Ablation studies for \model{}~model variations trained for 560k steps (32-48 batch size, 0.2 epochs) with 16 Volta 32 GB GPUs. Training time is in GPU days and the validation set is a 30k random sample of the OC20 ID Validation set.}
    \label{tab:comp-ablation}
    \vspace{-0.35cm}
\end{table*}
 
\begin{figure*}[t]
    \centering
    \includegraphics[width=\linewidth]{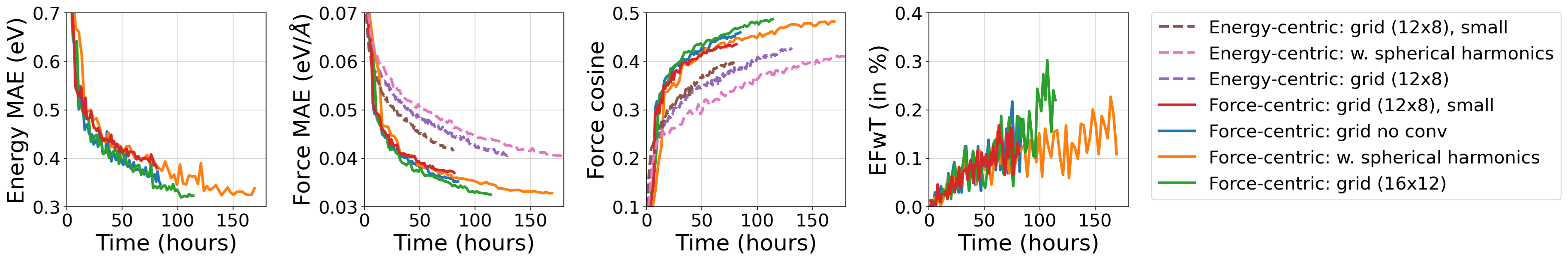}
    \caption{Performance of SpinConv ablations on OC20 Val ID $30k$ (Table~\ref{tab:comp-ablation}).
    All models trained for $560k$ steps and plotted against wall-clock training time.
    Note force-centric models and grid-based approaches converge more quickly than energy-centric models
    and those using spherical harmonics.}
    \label{fig:ablationplots}
\end{figure*}

In Table \ref{tab:comp-splits} we examine the performance of \model~across different test splits. Note that the energy prediction of \model~is signficantly better than SchNet or DimeNet++. Across all models the accuracy for the in domain split are highest and decline for the three Out of Domain (OOD Adsorbate, OOD Catalyst, OOD Both) splits. \model~outperforms all models on each of the different domain splits. When comparing energy-centric approaches trained with both force and energy losses (bottom rows), the \model~model does significantly better at predicting both. In fact, the energy-centric approach trained on forces and energy outperforms the DimeNet++ \cite{klicpera_dimenetpp_2020} model when trained on only energy, or energy and forces.

We examine variations of the \model~model in Table \ref{tab:comp-ablation} and Figure \ref{fig:ablationplots} through ablation studies. We trained three variants of the energy-centric model and four variants of the force-centric model. The grid-based and spherical harmonic approaches produced similar accuracies. However, the grid-based approach was significantly faster to train, so it was used in the remaining experiments. Smaller models lead to reduced performance on the OC20 dataset, but we found for smaller datasets such as MD17 or QM9 smaller model sizes can be beneficial to avoid overfitting. Finally, we test the impact of not performing the convolution (no conv) and only applying the filter at a single rotation. Rotation invariance was maintained by orienting the filter based on the mean angle of the neighboring atoms weighted by distance. The result of not performing the convolution is significantly reduced accuracy. However, its faster training time may make it suitable for some applications. 

Finally, for the force-centric \model~model we explore results when varying the number of random rotations used in the force block. The force MAE when using a single random rotation is 0.0276 and improves slightly to 0.0270 when using 5 random rotations. Increasing the number of rotations beyond 5 leads to negligible gains. The standard deviation of the force estimates at different random rotations is 0.004 eV/\AA. This is equal to $15\%$ of the force MAE, which indicates the amount of error due to the model not being strictly rotation equivariant is small relative to the overall error of the model.

\setlength{\tabcolsep}{3pt}
\begin{table*}[t]
    \begin{center}
        \resizebox{\textwidth}{!}{
            \begin{tabular}{l l cccc | cccc  }
            \midrule
            & & \multicolumn{4}{c|}{Energy MAE [eV] $\downarrow$} & \multicolumn{4}{c}{EwT $\uparrow$}  \\
            \cmidrule(l{4pt}r{4pt}){3-6}
            \cmidrule(l{4pt}r{4pt}){7-10}
        Model & Approach & ID &  OOD Ads & OOD Cat & OOD Both & ID &  OOD Ads & OOD Cat & OOD Both \\
            \midrule
                Median baseline & -
                    & $1.7499$ & $1.8793$ & $1.7090$ & $1.6636$
                    & $0.71\%$ & $0.72\%$ & $0.89\%$ & $0.74\%$ \\
                \midrule
                CGCNN~\cite{xie2018crystal} & Direct
                    & $0.6149$ & $0.9155$ & $0.6219$ & $0.8511$
                    & $3.40\%$ & $1.93\%$ & $3.10\%$ & $2.00\%$  \\
                SchNet~\cite{schutt2017schnet} & Direct
                    & $0.6387$ & $0.7342$ & $0.6616$ & $0.7037$
                    & $2.96\%$ & $2.33\%$ & $2.94\%$ & $2.21\%$  \\
                % DimeNet~\cite{klicpera2020directional} & Direct
                %     & $0.5974$ & $0.7568$ & $0.6119$ & $0.7097$
                %     & $3.85\%$ & $2.00\%$ & $3.32\%$ & $1.94\%$   \\
                DimeNet${+}{+}$~\cite{klicpera2020directional} & Direct
                    & $0.5620$ & $0.7252$ & $0.5756$ & $0.6613$
                    & $4.25\%$ & $2.07\%$ & $4.10\%$ & $2.41\%$ \\
                SpinConv & Direct
                    & 0.5583   &  0.7230   & 0.5687 & 0.6738
                    & 4.08\%   &  2.26\%   & 3.82\% & 2.33\%\\
                \midrule
                % SchNet~\cite{schutt2017schnet} & Relaxation
                %     & $1.8630$ & $1.9351$ & $2.1367$ & $2.0835$
                %     & $0.40\%$ & $0.80\%$ & $0.39\%$ & $0.39\%$ \\
                % SchNet~\cite{schutt2017schnet} -- force-only + energy-only & Relaxation
                %     & $1.6643$ & $1.6948$ & $1.9577$ & $1.8270$
                %     & $0.59\%$ & $0.40\%$ & $0.40\%$ & $0.59\%$ \\
                % DimeNet~\cite{klicpera2020directional} & Relaxation
                    % & $1.6199$ & $1.7281$ & $1.5812$ & $1.3781$
                    % & $1.21\%$ & $0.72\%$ & $1.38\%$ & $0.81\%$ \\
                DimeNet${+}{+}$ & Relaxation
                    & $0.6908$ & $0.6842$ & $0.7027$ & $0.6834$
                    & $4.25\%$ & $3.36\%$ & $3.76\%$ & $3.52\%$ \\
                DimeNet${+}{+}$ -- force-only + energy-only & Relaxation
                    & $0.5124$ & $0.5744$ & $0.5935$ & $0.6126$
                    & $6.12\%$ & $4.29\%$ & $5.07\%$ & $3.85\%$ \\
                DimeNet${+}{+}$ -- large force-only + energy-only & Relaxation
                    & $0.5034$ & $0.5430$ & $0.5789$ & $0.6113$
                    & $6.57\%$ & $4.34\%$ & $5.09\%$ & $3.93\%$ \\
                SpinConv (force-centric) & Relaxation
                    & $\textbf{0.4235}$ & $\textbf{0.4415}$ & $\textbf{0.4572}$ & $\textbf{0.4245}$
                    & $\textbf{9.37\%}$ & $\textbf{6.75\%}$ & $\textbf{8.49\%}$ & $\textbf{6.76\%}$\\
            \bottomrule
            \end{tabular}}
    \end{center}
    \caption{Initial Structure to Relaxed Energy (IS2RE) results on the OC20 test split as evaluated by the Energy MAE (eV) and Energy within Threshold (EwT) \cite{OC20} (see OC20 discussion board). Comparisons made for the direct and relaxation approaches using various models.}
    \label{tab:is2re}
\end{table*}

\begin{table*}[]
  \centering
    \resizebox{\columnwidth}{!}{
    \begin{tabular}{lrrrrrr|rrrrr}
      \toprule
        \mr{2}{\textbf{Model}}  & \textbf{Inference} & \mc{5}{c}{\textbf{AFbT (\%)} $\uparrow$} & \mc{5}{c}{\textbf{ADwT (\%)} $\uparrow$} \\
        & \textbf{time} $\downarrow$ &  ID & OOD Ads. & OOD Cat. & OOD Both & {\bf Average} &  ID & OOD Ads. & OOD Cat. & OOD Both & {\bf Average}  \\
      \midrule
        SchNet~\cite{schutt2017schnet} & 54.1h & 5.28 & 2.82 & 2.62 & 2.73 & 3.36 & 32.49 & 28.59 & 30.99 & 35.08 & 31.79 \\
        DimeNet++~\cite{klicpera_dimenetpp_2020} & 407.6h & 17.52 & 14.67 & 14.32 & 14.43 & 15.23 & 48.76 &  45.19 & 48.59 & 53.14 & 48.92 \\
        DimeNet++-large~\cite{klicpera_dimenetpp_2020} & 814.6h & \textbf{25.65} & \textbf{20.73} & \textbf{20.24} & \textbf{20.67} & \textbf{21.82} & 52.45 & 48.47 & 50.99 & 54.82 & 51.68 \\
        ForceNet~\cite{hu2021forcenet} & 75.1h & 10.75 & 7.74 & 7.54 & 7.78 & 8.45 & 46.83 & 41.26 & 46.45 & 49.60 & 46.04   \\
        ForceNet-large~\cite{hu2021forcenet} & 186.9h & 14.77 & 12.23 & 12.16 & 11.46 & 12.66 & 50.59 & 45.16 & 49.80 & 52.94 & 49.62   \\
        \midrule
        % \bf \model{} (energy-centric) \\
        \bf \model{} (force-centric) & 263.2h & 21.10 & 15.70 & 15.86 & 14.01 & 16.67
            & \textbf{53.68} & \textbf{48.87} & \textbf{53.92} & \textbf{58.03} & \textbf{53.62} \\  
      \bottomrule
    \end{tabular}}
    \caption{Relaxed structure from initial structure (IS2RS) results on
      the OC20 test split, as evaluated by Average Distance within Threshold (ADwT)
      and Average Forces below Threshold (AFbT). All values in percentages, higher is better.
      Results computed via the OCP evaluation server. Inference times are total across the $4$ splits.}
    \label{tab:results-is2rs}
\end{table*}

\subsubsection{Initial Structure to Relaxed Energy (IS2RE)}
The Initial Structure to Relaxed Energy (IS2RE) task takes an initial atomic structure and attempts to predict the energy of the structure after it has been relaxed. Two approaches may be taken to address this problem, the direct and relaxation approaches \cite{OC20}. The direct treats the task as a standard regression problem and directly estimates the relaxed energy from the initial structure. The relaxation approach computes the relaxed structure using the ML predicted forces to update the atom positions. Next, given the ML relaxed structure the energy is estimated. We show results for both approaches in the OC20 dataset using \model~in Table \ref{tab:is2re}. 

The results of the \model~model significantly outperform all previous approaches using the relaxation approach for both energy MAE and Energy within Threshold (EwT) metrics. DimeNet++ also shows improved results for the relaxation approach with the best approach using two models; DimeNet++-large for force estimation and DimeNet++ (energy-only) for the energy estimation. Note in contrast to other approaches, \model~shows good results across all test splits, including those with out of domain adsorbates and catalysts. Using the direct approach, \model~is comparable to DimeNet++'s direct approach.

\subsubsection{Initial Structure to Relaxed Structure (IS2RS)}

Our final results on the OC20 dataset are on the IS2RS task where predicted forces are used to relax an atom structure to a local energy minimum. The is performed by iteratively estimating the forces that are in turn used to update the atoms positions. This process is repeated until convergence or 200 iterations. Results are shown in Table~\ref{tab:results-is2rs}. The suggested metrics are  Average Distance within Threshold (ADwT) metric, which measures whether the atom positions are close to those found using DFT
and Average Forces below Threshold (AFbT), which measures whether a true energy minimum was found (i.e., forces are close to zero). On the ADwT metric, \model~outperforms other approaches ($53.62\%$ averaged across splits).
On the AFbT metric, DimeNet++-large outperforms \model~($21.82\%$~\vs $16.67\%$),
but is more than ${\sim}3$x slower ($814.6$h~\vs $263.2$h) during inference. \model~outperforms all other models.

\begin{table*}[t]
    \centering
    \resizebox{0.7\columnwidth}{!}{   
    
    \renewcommand{\arraystretch}{1.0}
    \setlength{\tabcolsep}{5pt}
    \resizebox{0.97\linewidth}{!}{
    \begin{tabular}{lccc|ccc}
      \toprule
        Molecule & GDML &  PhysNet &  PhysNet-ens5 & SchNet & DimeNet* & \model \\
        \midrule
        Aspirin         & 0.02 & 0.06 & 0.04 & 0.33 & 0.09 & \textbf{0.07} \\ 
        Benzene         & 0.24 & 0.15 & 0.14 & 0.17 & \textbf{0.15} & 0.17 \\
        Ethanol         & 0.09 & 0.03 & 0.02 & 0.05 & 0.03 & \textbf{0.02} \\
        Malonaldehyde   & 0.09 & 0.04 & 0.03 & 0.08 & \textbf{0.04} & \textbf{0.04} \\
        Naphthalene     & 0.03 & 0.04 & 0.03 & 0.11 & 0.06 & \textbf{0.04} \\
        Salicylic       & 0.03 & 0.04 & 0.03 & 0.19 & 0.09 & \textbf{0.05} \\
        Toluene         & 0.05 & 0.03 & 0.03 & 0.09 & 0.05 & \textbf{0.03} \\
        Uracil          & 0.03 & 0.03 & 0.03 & 0.11 & 0.04 & \textbf{0.03} \\
        \midrule
        Mean & 0.073 & 0.053 & 0.044 & 0.141 & 0.069 & \textbf{0.058} \\
      \bottomrule
    \end{tabular}
    }}
    \caption{Forces MAE (kcal/mol\AA) on MD17 for models trained using 50k samples. Best results for models not using domain specific information are in bold. {\footnotesize *The DimeNet results were trained in-house as the original authors did not use the 50k dataset. DimeNet was found to outperform DimeNet++ on this task.} 
    }
    \vspace{-0.35cm}
    \label{tab:comp-md17}
\end{table*}

\begin{table*}[]
    \centering
    \renewcommand{\arraystretch}{1.0}
    \resizebox{0.80\linewidth}{!}{
    \begin{tabular}{lcccccccccccc}
        \toprule
        Task & $\alpha$ & $\Delta \epsilon$ & $\epsilon_{\text{HOMO}}$ & $\epsilon_{\text{LUMO}}$ & $\mu$ & $C_{\nu}$ & G &
        H & R$^{2}$ & U & U$_{0}$ & ZPVE \\
        Units & bohr$^{3}$ & meV & meV & meV & D & cal/mol K & meV & 
        meV & bohr$^{3}$ & meV & meV & meV \\
        \midrule
        NMP \cite{gilmer2017neural} &       .092 & 69 & 43 & 38 & .030 & .040 & 19 & 17 & .180 & 20 & 20 & 1.50 \\
        Schnet \cite{schutt2017schnet} &    .235 & 63 & 41 & 34 & .033 & .033 & 14 & 14 & \textbf{.073} & 19 & 14 & 1.70 \\
        Cormorant \cite{anderson2019cormorant} & .085 & 61 & 34 & 38 & .038 & .026 & 20 & 21 & .961 & 21 & 22 & 2.03 \\
        L1Net \cite{miller2020relevance} &     .088 & 68 & 46 & 35 & .043 & .031 & 14 & 14 & .354 & 14 & 13 & 1.56 \\
        LieConv \cite{finzi2020generalizing} &   .084 & 49 & 30 & 25 & .032 & .038 & 22 & 24 & .800 & 19 & 19 & 2.28 \\
        TFN \cite{thomas2018tensor} &       .223 & 58 & 40 & 38 & .064 & .101 & - & - & - & - & - & - \\
        SE(3)-Tr. \cite{fuchs2020se} & .142 & 53 & 35 & 33 & .051 & .054 & - & - & - & - & - & - \\
        EGNN \cite{satorras2021n} &      .071 & 48 & 29 & 25 & .029 & .031 & 12 & 12 & .106 & 12 & 11 & 1.55\\
        DimeNet++ \cite{klicpera_dimenetpp_2020} & \textbf{.044} & 33 & 25 & 20 & .030 & .023 & \textbf{8} & 7 & .331 & \textbf{6} & \textbf{6} & 1.21 \\
        SphereNet \cite{liu2021spherical} & .047 & \textbf{32} & \textbf{24} & \textbf{19} & \textbf{.027} & \textbf{.022} & \textbf{8} & \textbf{6} & .292 & 7 & \textbf{6} & \textbf{1.12} \\
        \midrule
        \textbf{\model{}} & .058 & 47 & 26 & 22 & \textbf{.027} & .028 & 12 & 12 & .156 & 12 & 12 & 1.50\\
        % \midrule
        \bottomrule
    \end{tabular}
    }
    
    \caption{Mean absolute error results for QM9 dataset \cite{ramakrishnan2014quantum} on 12 properties for small molecules.}
    \label{tab:comp-qm9}
\end{table*}

\subsection{MD17}

The MD17 dataset \cite{chmiela2017machine,chmiela2018towards} contains molecular dynamic simulations for eight small molecules. Two training datasets are commonly used, one containing 1k examples and another containing 50k examples. We found the 1k training dataset to be too small for the \model~model, and may be more appropriate for approaches that incorporate prior chemistry knowledge, such as hand-coded features or force fields \cite{chmiela2017machine,unke2019physnet}. The 50k dataset provides significantly more training data, but the remaining validation and test data are highly similar to those found in training, and may not guarantee independent samples in the test set\cite{christensen2020role}. Nevertheless, we report results on MD17 for comparison to prior work on the molecular dynamics task. Research in this domain would greatly benefit from the generation of a larger dataset. 

Results are shown in Table \ref{tab:comp-md17}. \model~is on par or better for 7 of the 8 molecules when compared to DimeNet \cite{klicpera2020directional}. Both \model~and DimeNet perform well with respect to the GDML \cite{chmiela2017machine} and PhysNet \cite{unke2019physnet} models that take advantage of domain-specific information. Given the smaller dataset size, the \model~model uses a reduced 8x8 grid-based spherical representation. Other model parameters are the same as previously described.

\subsection{QM9}

Our final set of results are on the popular QM9 dataset \cite{ramakrishnan2014quantum} that tests the prediction of numerous properties for small molecules. While the \model~model was designed to estimate energies and per-atom forces, we may use the same model to predict other proprieties. Results are shown in Table \ref{tab:comp-qm9} on a random test split for an energy-centric 8x8 grid-based \model~model. The results of DimeNet++ and the recent SphereNet\cite{liu2021spherical} outperform those of others. However, DimeNet++, SphereNet and \model~perform well with respect to other approaches across many properties.

% We compare to NMP \cite{gilmer2017neural}, Schnet \cite{schutt2017schnet}, Cormorant \cite{anderson2019cormorant}, L1Net \cite{miller2020relevance}, LieConv \cite{finzi2020generalizing}, TFN \cite{thomas2018tensor}, SE(3)-Tr \cite{fuchs2020se} and EGNN \cite{satorras2021n}.

\section{Related work}
\label{sec:related}
A common approach to estimating molecular and atomic properties is the use of GNNs  \cite{schutt2017quantum,gilmer2017neural,jorgensen2018neural,schutt2017schnet,schutt2018schnet,xie2018crystal,qiao2020orbnet,klicpera2020directional} where nodes represent atoms and edges connect neighboring atoms. One of the first approaches for force estimation was SchNet \cite{schutt2017schnet}, which computed forces using only the distance between atoms without the use of angular information. Unlike previous approaches that used discrete distance filters \cite{xie2018crystal}, SchNet proposed the used of differentiable edge filters. This enabled the construction of an energy-conserving model for molecular dynamics that estimates forces by taking the negative gradient of the energy with respect to the atom positions \cite{chmiela2017machine}. DimeNet extended this approach to also represent the angular information between triplets of atoms \cite{klicpera2020directional,klicpera_dimenetpp_2020}. The more recent SphereNet further extends this by capturing dihedral angles \cite{liu2021spherical}. \model~is able to model relative angular relationships between all neighboring atoms, and not just triplets of atoms, due to the use of the spin convolutional filter. In parallel to invariant models, rotational equivariant networks are explored in depth by \cite{weiler20183d,batzner2021se,anderson2019cormorant,thomas2018tensor,satorras2021n}. This was accomplished by decoupling the network-fed invariant information (distance), from the equivariant information (distance vector), followed by the careful combination via tensor products. The energy-centric \model~ model is invariant to rotations due to the use of global pooling after the spin convolution. The final force block of the force-centric model is not strictly rotation equivariant, but is encouraged to learn rotation equivariance during training.

Another approach to force estimation is to directly regress the forces as an output of the network. This doesn't enforce energy conservation or rotational equivariance, but as shown by ForceNet \cite{hu2021forcenet}, such models can still produce accurate force estimates. 

Numerous approaches incorporate more domain specific information into machine learning models. These include GDML \cite{chmiela2017machine} and PhysNet \cite{unke2019physnet} that use handcrafted features and force-fields respectively. OrbNet \cite{qiao2020orbnet} is a hybrid approach that utilizes proprietary orbital features that improves accuracy while achieving significant efficiency gains over DFT. While these approaches can lead to improved accuracy, they typically result in increased computational expense over ML models.

\section{Discussion}
\label{sec:discussion}
While the \model~model demonstrates improved performance, it still has significant limitations. Most notable is the force and energy estimates are still significantly lower than desired for practical applications. Further research is needed to improve accuracies, so that machine learning models can be widely adopted. Currently, the \model~model does not take advantage of domain specific information. Results could be significantly improved, especially for smaller datasets (e.g., MD17 1k), if more domain information was integrated into the model \cite{chmiela2017machine,unke2019physnet,qiao2020orbnet}. The use of the spin convolution becomes increasingly expensive as the size of the filter increases, since the number of convolutions is equal to the longitudinal dimension of the filter. If filters of higher resolution are needed, more computationally efficient approaches may be required.

In conclusion, we propose the \model~model that effectively captures the relative angular information of neighboring atoms, while maintaining the invariance of the energy estimation with respect to system rotations. This is enabled by utilizing a spin convolution over a spherical representation in a per-edge local reference frame, followed by global pooling. Two model variants are proposed based on whether energy conservation is enforced. Results demonstrate state-of-the-art results on the OC20 dataset, and strong results on both the MD17 and QM9 datasets.

\section{Societal Impact}

This work is motivated by the problems we face due to climate change \cite{zitnick2020introduction}, many of which require innovative solutions to reduce energy usage and replace traditional chemical feedstocks with renewable alternatives. For example, one of the most energy intensive chemical processes is the development of new electrochemical catalysts for ammonia fertilizer production that helped to feed the world's growing population during the 20th century \cite{hager2009alchemy}. This is also an illustrative example of possible unintended consequences as  advancements in chemistry and materials may be used for numerous purposes. As ammonia fertilization increased in use, its overuse in today's farming has led to ocean ``dead zones'' and its production is very carbon intensive. Knowledge and techniques used to create ammonia were also transferred to the creation of explosives during wartime.  We hope to steer the use of ML for atomic simulations to societally-beneficial uses by training and testing our approaches on datasets, such as OC20, that were  specifically designed to address chemical reactions useful for addressing climate change.  

\bibliography{reference}

\end{document}